\documentclass[10pt,twocolumn,letterpaper]{article}
\usepackage{cvpr}
\usepackage{cite}
\usepackage{color}
\usepackage[pdftex]{graphicx}
\graphicspath{{figures/}}
\DeclareGraphicsExtensions{.pdf,.jpeg,.png,.eps}
\usepackage[cmex10]{amsmath}
\usepackage{algorithmicx,algorithm}
\usepackage{tabularx}
\usepackage{longtable}
\usepackage{pbox}
\usepackage{array}
\usepackage{threeparttable}
\usepackage{multirow}
\usepackage{url}

\usepackage{amsmath,amssymb}
\usepackage{latexsym}
\usepackage{CJK}
\usepackage{graphicx}

\newcommand{\tabincell}[2]{\begin{tabular}{@{}#1@{}}#2\end{tabular}}
\newcommand{\mytabspace}{\vspace*{-1.0em}}
\newcommand{\myfigspace}{\vspace*{-1.0em}}

\usepackage[pagebackref=true,breaklinks=true,letterpaper=true,colorlinks,bookmarks=false]{hyperref}

 \cvprfinalcopy 

\def\cvprPaperID{****} 

\ifcvprfinal\fi
\begin{document}

\title{Learning Compact Appearance Representation for Video-based Person Re-Identification}

\author{Wei Zhang$^1$, Shengnan Hu$^1$, Kan Liu$^1$, Zhengjun Zha$^2$\\
$^1$ School of Control Science and Engineering, Shandong University, China\\
$^2$ University of Science and Technology of China \\
{\tt\small davidzhangsdu@gmail.com}
}
\maketitle

\begin{abstract}
This paper presents a novel approach for video-based person re-identification using multiple Convolutional Neural Networks (CNNs). Unlike previous work, we intend to extract a compact yet discriminative appearance representation from several frames rather than the whole sequence. Specifically, given a video, the representative frames are selected based on the walking profile of consecutive frames. A multiple CNN architecture incorporated with feature pooling is proposed to learn and compile the features of the selected representative frames into a compact description about the pedestrian for identification. Experiments are conducted on benchmark datasets to demonstrate the superiority of the proposed method over existing person re-identification approaches.

\end{abstract}

\section{Introduction}

Person re-identification(re-id) has been widespread concerned recently, as this issue underpins various critical applications such as video surveillance, pedestrian tracking and searching. Given a target person appearing in a surveillance camera, a re-id system generally aims to identify it in the other cameras through the whole camera-network, i.e., determining whether instances captured by different cameras belong to the same person. However, due to the influence of cluttered background, occlusions and viewpoint variations across camera views, this task is quite challenging.


A re-id system may have an image or a video as input for feature extraction. Since only limited information can be exploited from a single image, it is difficult to overcome the occlusion, camera-view and pose variation problems and to capture the varying appearance of a pedestrian performing different action primitives.  Thus it is better to deal with the video-based re-id problem, as videos inherently contain more temporal information of the moving person than an independent image, not to mention in many practical applications the input are videos to begin with. Besides, video is a sequence of images, so spatial and temporal cues are more abundant in a video than in a image, which can facilitate extracting more features.
\begin{figure}[htbp]
\begin{center}
\includegraphics[width=0.8\linewidth]{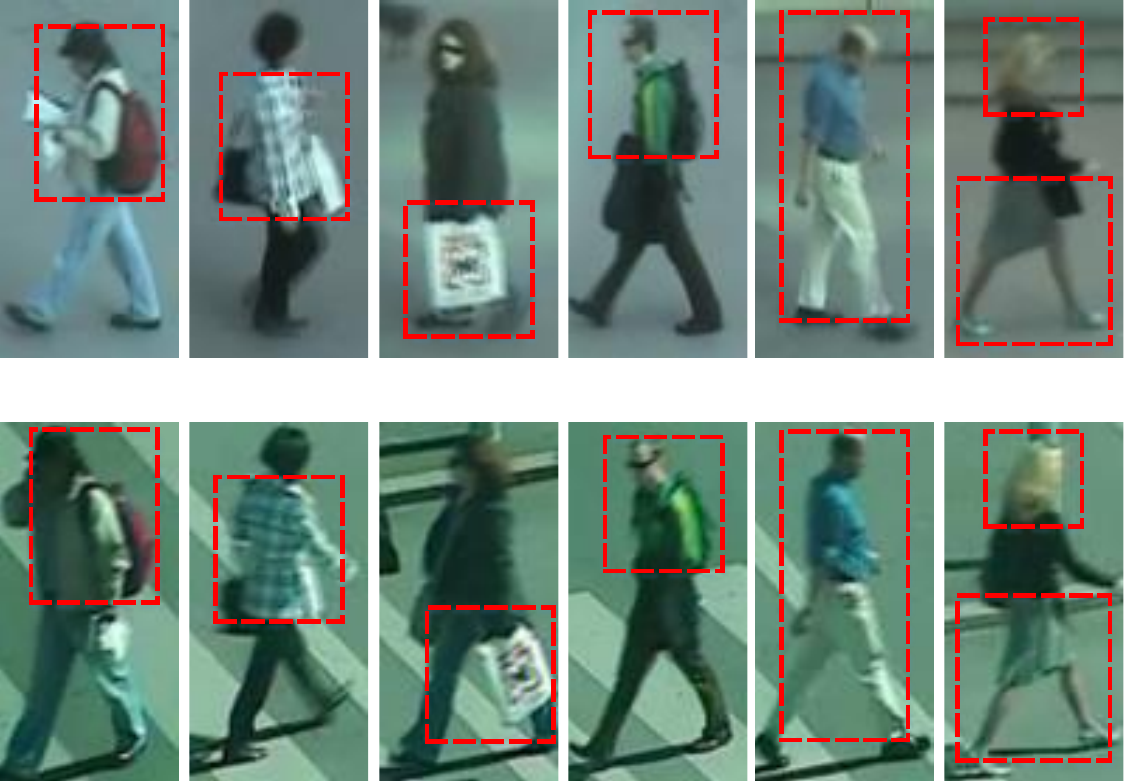}
\caption{Salient appearance in person re-id.}\label{fig:appearance}
\end{center}
\end{figure}

\begin{figure*}[htbp]
\begin{center}
\includegraphics[width=\linewidth]{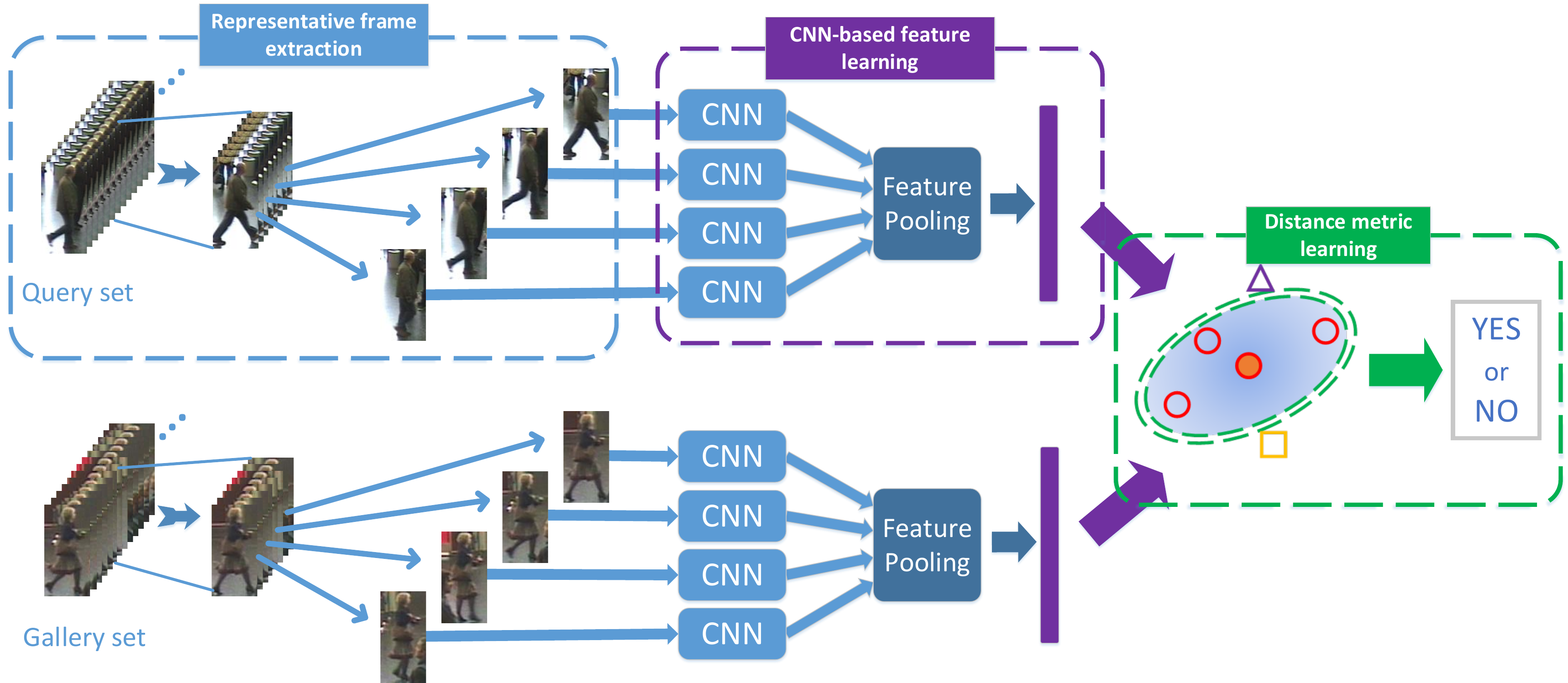}
\caption{An overview of the proposed video-based re-id framework.}\label{fig:overview}
\end{center}
\end{figure*}

In spite of the rich space-time information provided by a video sequence, more challenges come along. So far, only a few video-based methods have been presented \cite{WangECCV14}, \cite{liu2015spatio}, \cite{mclaughlinrecurrent}. Most of them focus on investigating the temporal information related to person's motion, such as their gait, and perhaps even the patterns of how their bodies and clothes move. Although such movement is one type of behavioral biometrics, it is unfortunate that a large number of persons share similarity in walking manners and related behavior \cite{you2016top} \cite{zheng2016mars}. Moreover, since gait is considered a biometric that is not affected by the appearance of a person, most approaches tried to exploit it by working with silhouettes, which are difficult to extract, especially from surveillance data with cluttered background and occlusions \cite{liu2015spatio}. Besides, time-series analysis usually requires extracting information at different timescales \cite{mclaughlinrecurrent}. In the person re-id problem, gait information often exists in short time, thus the information provided by movement descriptors is limited. In some cases, it is even harder to distinguish the video representations of different identities than the still-image appearance \cite{you2016top}.


%

Unlike previous work, in this paper we intend to extract a compact appearance representation from several representative frames rather than the whole frames for video-based re-id. Compared to the temporal-based methods, the proposed appearance model works more similarly to human visual system. Because the visual perception studies on appearance (e.g., color, texture) and motion stimuli have shown that the pattern detection thresholds are much lower than the motion detection thresholds \cite{seiffert1999position} \cite{lindsey1990motion} \cite{derrington1993discriminating}. Hence, human performs better at identifying the appearance of human body or belongings than the manners of how a person walks. In most cases, people can be distinguished more easily from appearance such as clothes and bags on their shoulders than from gait and pose which are generally similar among different persons \cite{NakajimaPR03}, as shown in Fig. \ref{fig:overview}. So, videos are highly redundant and it is unnecessary to incorporate all frames for person re-id. Our study shows that several typical frames with appropriate feature extraction can offer competitive or even better identification performance.



More specifically, given a walking sequence, we first split it into a couple of segments corresponding to different action primitives of a walking cycle. The most representative frames are selected from a walking cycle by exploiting the local maxima and minima of the Flow Energy Profile (FEP) signal \cite{WangECCV14}. For each frame, we propose a CNN to learn feature based on person's joint appearance information. Since different frames may have different discriminative features for recognition, by introducing an appearance-pooling layer, the salient appearance features of multiple frames are preserved to form a discriminative feature descriptor for the whole video sequence. The central point of our algorithm lies in the exploration of the key appearance information of a video, contrary to the conventional methods like \cite{mclaughlinrecurrent} and \cite{liu2015spatio}, which highly rely on accurate temporal information.

\section{Related work}
Person re-identification has been a hot topic in the computer vision community for decades \cite{ZhaoICCV13} \cite{XuICCV13} \cite{BakAVSS10a}. In general, the key is to generate discriminative signatures for pedestrian representation across different cameras. The most frequently used low-level features are color, texture, gradient, and the combination of them \cite{MaBMVC12} \cite{BedagkarPRL12} \cite{liu2012person}, extracted either from the whole body area or from the regions of interest.

Another popular way of synthesizing feature descriptors is through deep learning \cite{ouyang2016learning} \cite{chen2016deep}, which has shown great potential in various tasks of computer vision, such as object detection, image classification, face and pose recognition. In these areas, deep neural networks have largely replaced traditional computer vision pipelines based on hand-crafted features \cite{mclaughlinrecurrent}. As for the task of image-based person re-id, different CNNs have been used for learning a joint representation of images and similarity between image pairs or triplets directly from the pixels of the input images \cite{LiCVPR14} \cite{ding2015deep} \cite{xiao2016learning}.


Recently, the attention is moving to the video-based re-id problem and most efforts were spent on exploiting the temporal cues for the pedestrian modeling. Specifically, Wang et al. \cite{WangECCV14} employed the HOG3D \cite{KlaserBMVC08} as descriptor for action and activity recognition. Liu et al. \cite{liu2015spatio} developed a spatio-temporal alignment of video segments by tracking the gait information of pedestrians. Zheng et al. \cite{zheng2016mars} attempted to extract the motion by exploiting the HOD3G \cite{KlaserBMVC08} feature and the Gait Energy Image (GEI) \cite{man2006individual} feature. Some efforts even were spent on using hybrid tools such as RNN $+$ optical flow \cite{mclaughlinrecurrent} for temporal information extraction. However, as aforementioned, the temporal cues such as gait and motion are often unreliable from a walking sequence which is often short and of low quality in practical surveillance videos.

As an alternative, the proposed method is more like an image-based re-id algorithm. Following the human visual system, this work intends to solve the video re-id problem by pooling the distinctive features from several representative frames. To select the representative frames automatically, we do need some temporal cues to extract the walking cycle. Compared to the conventional temporal methods like optical flow, it is much simpler and does not need to be very accurate. As shown in Fig. \ref{fig:walkingcycle}, a rough approximate about the motion profile of consecutive frames is good enough. This could be also regarded as an implicit and more efficient way of using the temporal information, which may relieve the burdens of accurate motion or gait extraction in video re-id.

\section{Proposed method}
\label{sec:algorithm}

As illustrated in the Fig.~\ref{fig:overview}, our method proceeds in three steps: frame selection, feature pooling and identification. Given a video sequence, some representative frames are selected automatically based on the walking profile. Then each representative frame is processed by a CNN to extract reliable features. To compile all features into a compact yet informative description, a feature pooling layer is incorporated. Finally, we employ distance metric learning for identification, which maximizes the distance between features of different people and minimizes the distance of features of the same people.


\subsection{Representative frame extraction}
\label{sec:frame}

\begin{figure}[htbp]
\begin{center}
\includegraphics[width=\linewidth]{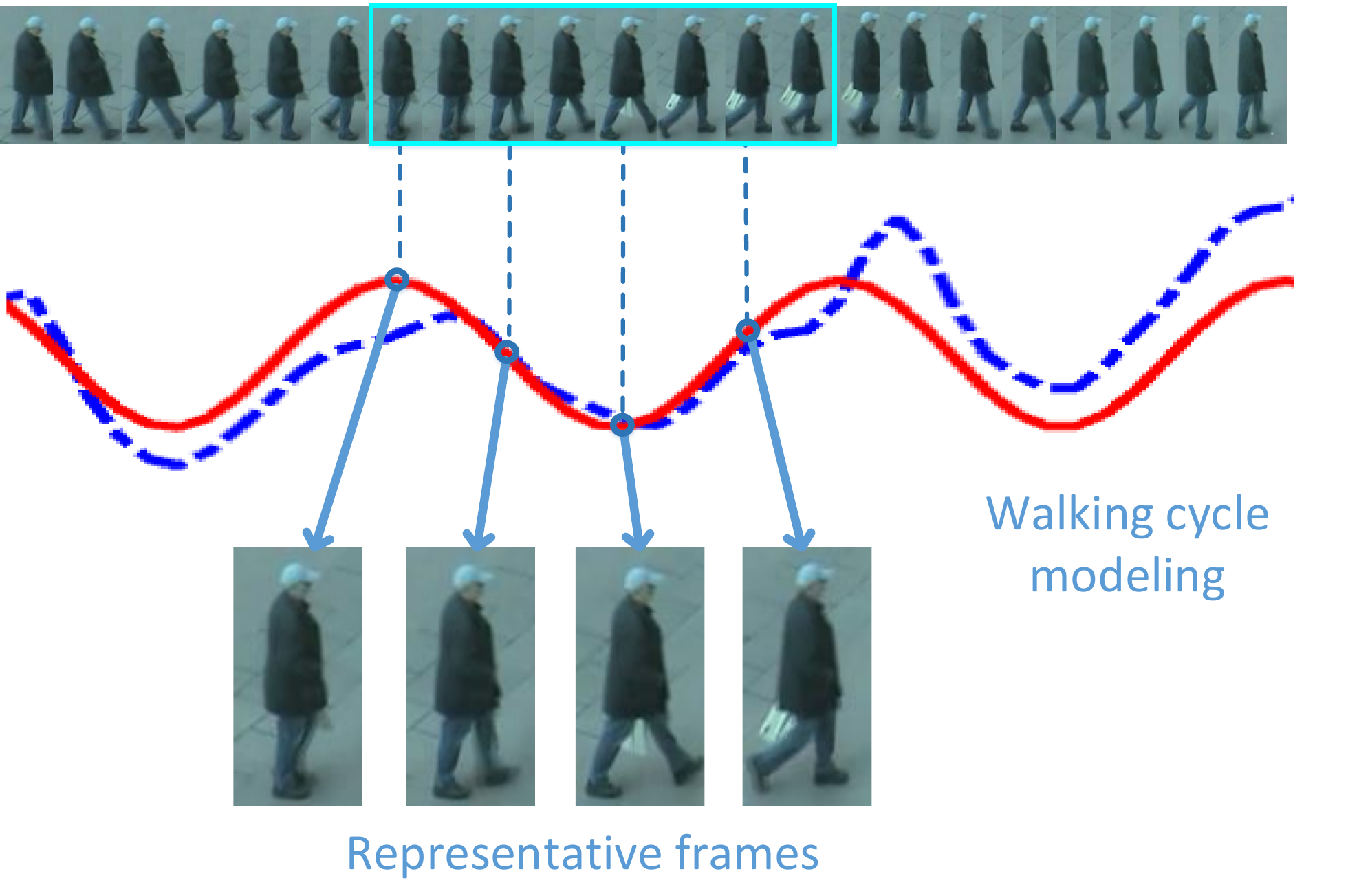}
\caption{Representative frame extraction. First row is a video sequence of a pedestrian and second row shows the related original FEP (blue curve) and the regulated FEP (red curve). The last row shows the four representative frames sampled based on the regulated FEP.}\label{fig:walkingcycle}
\end{center}
\end{figure}

To automatically select the most representative frames, we first extract the Flow Energy Profile (FEP) as proposed in \cite{WangECCV14}, which is a one dimensional signal denoted by $E$ that approximates the motion energy intensity profile of the consecutive frames in a video sequence. Ideally, the local maximum of $E$ corresponds to the postures when the person's two legs overlap, while at the local minimum the two legs are the farthest away. However, as shown in Fig.~\ref{fig:walkingcycle}, it can only provide a rough approximate about the walking circle as the estimation of FEP is sensitive to the noisy background and occlusions. Inspired by \cite{liu2015spatio}, the discrete Fourier transform is further employed to transform the FEP signal into the frequency domain, and the walking cycles can be better indicated by the dominant frequencies.

A full cycle of the walking action contains two consecutive sinusoid curves, one step from each leg. Since it is extremely difficult to distinguish between the two, each sinusoid curve of a single step is regarded as a walking cycle. Given a walking cycle, we can obtain the key frames corresponding to the different action primitives. As illustrated in Fig.~\ref{fig:walkingcycle}, the frames with the maximum FEP value and minimum FEP value are the best candidates for the representation of a walking cycle. The other frames can be sampled equally between the maximum and minimum of the circle. The studies in Table \ref{tab:views} show that four frames sampled from one circle give the best identification result. Adding more frames does not help, as most appearance information are already included.

\begin{figure*}[htbp]
\begin{center}
\includegraphics[width=\linewidth]{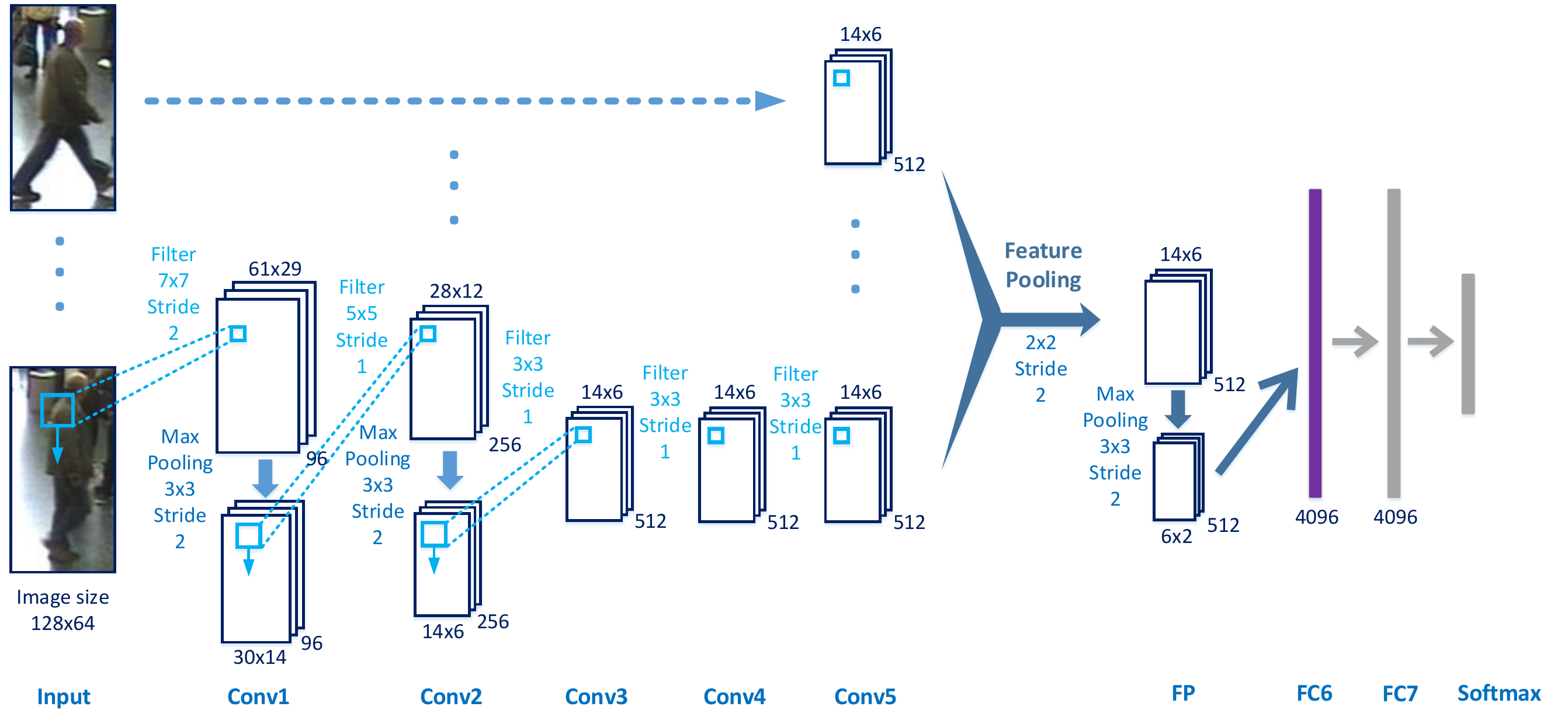}
\caption{Illustration of the proposed CNN architecture. The first five layers are convolutional layers and the last three layers are fully connected layers. The proposed feature pooling (FP) layer is between $Conv_5$ and $FC_6$.}\label{fig:cnn}
\end{center}
\end{figure*}
\subsection{CNN-based feature extraction}

\subsubsection{Network architecture}
The proposed network consists of five convolutional layers ($Conv_1,...,_5$) followed by two fully connected layers ($FC_6,_7$) and a softmax classification layer which is similar to the VGG-M network \cite{chatfield2014return}. The detailed structures are given in Fig.~\ref{fig:cnn}. To aggregate the features from the extracted representative frames into a single compact descriptor, a feature pooling layer is introduced to the network. Besides, the Rectified Linear Unit ($ReLU$) is employed as the neuron activation function. The Batch Normalization ($BN$) layers are used before each $ReLU$ layer, which can accelerate the convergence process and avoid manually tweaking the weights and bias \cite{xiao2016learning}.



The parameters of network are initialized from the pre-trained VGG-M model and then finetuned on the target training pedestrian sequence. At the training phase, the whole selected representative frames of each walking cycle are firstly rescaled to $128 \times 64 $, and then fed into the CNNs along with their corresponding label to train the network.

At the testing phase, the proposed network can be considered as a feature extractor using the CNN architectures. Specifically, each of the rescaled frame is first fed into the CNN to obtain its features with the convolutional layers. The learnt descriptors are then aggregated by a feature pooling layer and finally turns to be a 4096 dimensional representation at the fully connected layers. Note that, the features yielded at the $FC_6$ layer gave the best performance in experiments. So, the $FC_7$ and $Softmax$ layers are discarded after training. 

\begin{figure}[htbp]
\begin{center}
\includegraphics[width=\linewidth]{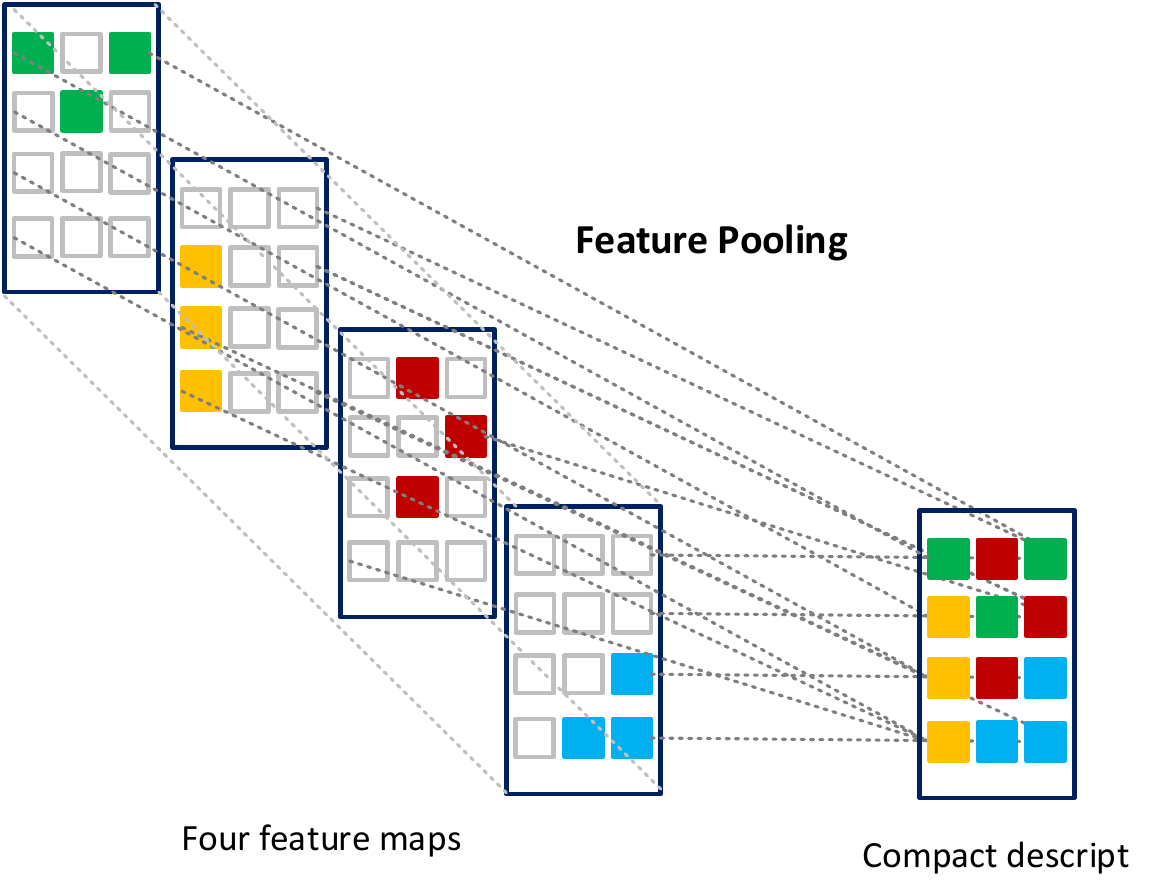}
\caption{Pooling salient features of multiple frames to form a single compact descriptor.}\label{fig:pooling}
\end{center}
\end{figure}

\subsubsection{Feature pooling}
\label{sec:pooling}
In this section, we focus on aggregating the key information from different views into a single, compact feature descriptor. After feeding the representative frames, the proposed CNN architecture will yield multiple feature maps as shown in Fig. \ref{fig:cnn}. Simply averaging these features is a straightforward way, but often leads to inferior performance \cite{su2015multi}. A feature pooling layer is added to the proposed CNNs. As shown in Table \ref{tab:pooling}, max pooling across the feature maps obtained from multiple CNNs produced the best re-id results.


Specifically, as illustrated in Fig.~\ref{fig:pooling}, although CNN is able to capture information from each frame, the discriminative appearance features of a pedestrian may appear in any frame, i.e., the desired discriminative features are scattered among different frames. However, by using the element-wise maximum operation among the feature maps, the strongest features from different views can be integrated to form a informative description about the pedestrian. Theoretically, this pooling layer can be inserted anywhere of the proposed network, yet the experimental results show that it performs best to be placed between the last convolutional layer and the first fully connected layer.

\subsection{Distance metric learning}
After feature extraction and pooling, to compare the final representation, we learn a metric on the training set using distance metric learning approaches.
Specifically, for each pedestrian representation $\mathbf{x}$ with $n_x$ feature vectors ($\mathbf{x}_i$) from the query set and representation $\mathbf{y}$ with $n_y$ feature vectors ($\mathbf{y}_j$) from the gallery set, the minimum distance of all the feature pairs $(\mathbf{x}_i,\mathbf{y}_j)$ is adopted as the distance $d$ between them as follows:

\begin{equation} \label{measure1}
    d_{min}(\mathbf{x}, \mathbf{y}) = \min_{i,j}\|\mathbf{x}_i - \mathbf{y}_j\|_2.
\end{equation}

An alternative is using the average of the minimum distance as the distance measurement between each feature pair as below:
\begin{equation} \label{measure2}
    d_{avg}(\mathbf{x}, \mathbf{y}) = \frac{\sum_i\min_j\|\mathbf{x}_i - \mathbf{y}_j\|_2}{2n_x} + \frac{\sum_j\min_i\|\mathbf{x}_i - \mathbf{y}_j\|_2}{2n_y}.
\end{equation}

Empirically, it is found that the latter measurement gives better performance. Besides, PCA is first performed to reduce the dimension of the original representation before distance metric learning and we choose the same reduced dimension as 100 in all of our experiments. More analysis and discussion about distance learning and dimension reduction can be found in Section \ref{sec:kissme}.

\section{Experimental results}

In this section, we conducted experiments on benchmark video re-identification datasets and made comparison between the proposed method and state-of-the-art approaches.

\subsection{Datasets and settings}

Experiments were conducted on three person re-id datasets: PRID 2011 dataset \cite{HirzerSCIA11}, iLIDS-VID dataset \cite{WangECCV14} and SDU-VID dataset \cite{liu2015spatio}. The PRID 2011 dataset includes 400 images sequences for 200 persons, captured by two non-overlapping cameras, and the average length of each sequence is 100. This dataset was captured in uncrowded outdoor scenes with relatively simple and clean background. The iLIDS-VID dataset contains 600 image sequences for 300 randomly sampled persons, with an average length of 73. This dataset was captured by two non-overlapping cameras in an airport hall under a multi-camera CCTV network. Subject to quite large illumination changes, occlusions, and viewpoints variations across camera views, this dataset is more challenging. The SDU-VID dataset \cite{liu2015spatio} contains 600 image sequences for 300 persons captured by two non-overlapping cameras. There are more image frames in each video sequence, and the average length is 130. This is also a challenging dataset due to the cluttered background, occlusions and viewpoint variations.

In our experiments, all datasets are randomly divided into training set and testing set by half, with no overlap between them. During testing, we consider the sequences from the first camera as the query set while the other one as the gallery set. For each walking cycle extracted from the video sequences, four representative frames are selected automatically as the inputs to four independent CNNs, which finally output a 4096-D descriptor for the whole walking cycle. Since different video sequences may contain different numbers of walking cycles, for each sequences we may extract a different number of feature descriptors. We use all of them as query or gallery descriptors and learn a metric to determine the distance between two sets of descriptors extracted from two sequences. The widely used Cumulative Matching Characteristics (CMC) curve is employed for quantitative measurement. All tests will be repeated 10 times and the average rates is reported to ensure statistically reliable evaluation.

\subsection{Results of feature learning}
\begin{figure}[htbp]
\begin{center}
\includegraphics[width=\linewidth]{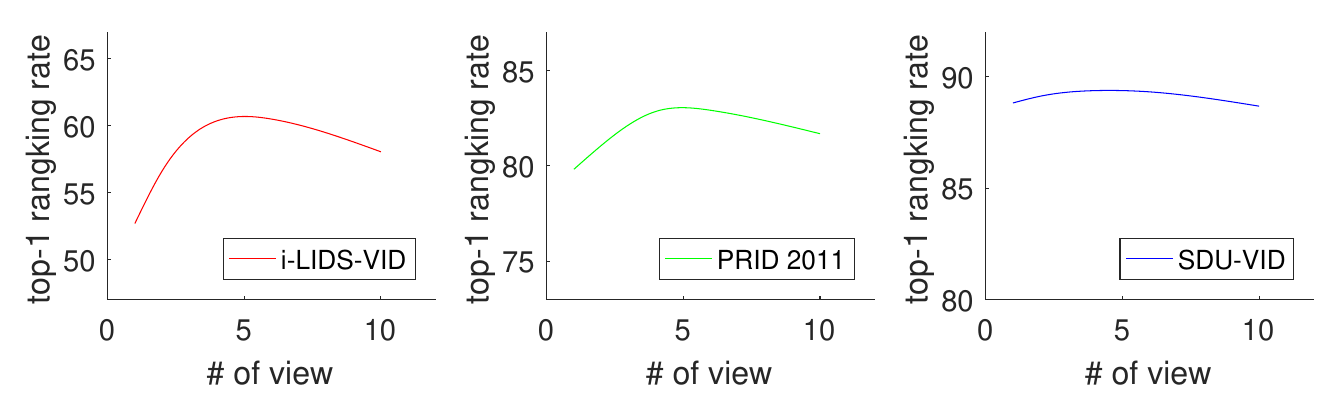}
\caption{Effect of the number of frames sampled in a walking circle for re-id. }\label{fig:viewN}
\end{center}
\myfigspace
\end{figure}

\begin{table}[htb]
  \centering
  \caption{Re-id performance with different number of input frames}
  \label{tab:views}
    \begin{tabular}{c|cc|cc|cc} \hline
    Dataset & \multicolumn{2}{c|}{iLIDS} & \multicolumn{2}{c|}{PRID} & \multicolumn{2}{c}{SDU} \\ \hline
    Frames & R-1 & R-5  & R-1 & R-5  & R-1 & R-5 \\ \hline

1          & 50.9 &80.4 &79.7 &91.7 &88.7 &94.7 \\
2          & 58.9 &81.8 &81.0 &92.3 &{\bf89.3} &94.7 \\
4          & {\bf60.2} &85.1 &{\bf83.3} &{\bf93.3} & {\bf89.3} &{\bf95.3}\\
6          & {\bf60.2} &{\bf85.3} &82.7 &{\bf93.3} &{\bf89.3} &{\bf95.3} \\
10         & 58.0 &84.7 &81.7 &92.3   &88.7 &{\bf95.3}  \\
    \hline
    \end{tabular}
\end{table}

\begin{table}[htb]
  \centering
  \caption{Comparison to different frame selection methods}
  \label{tab:frames}
  \begin{threeparttable}
    \begin{tabular}{c|c|c|c} \hline
    Dataset & \multicolumn{1}{c|}{iLIDS} & \multicolumn{1}{c|}{PRID} & \multicolumn{1}{c}{SDU} \\ \hline
    Frame selection method  & R-1  & R-1   & R-1  \\ \hline
Random selection-1\tnote{1}  &44.4 &77.0  &86.0   \\
Random selection-2\tnote{2}  &48.7 &77.0  &87.8  \\
All frames   &58.0 &83.0  &88.7  \\
Proposed sampling method  &{\bf60.2}  &{\bf83.3}  &{\bf89.3}   \\
    \hline
    \end{tabular}
 \begin{tablenotes}
        \footnotesize
\item[1] Randomly sample $K$ frames among the whole sequence. $K$ is determined automatically as stated in Section \ref{sec:frame}.
\item[2] Equally divide the video into $K$ segments and sample one frame from each.
      \end{tablenotes}
\end{threeparttable}
\end{table}




\paragraph{Representative frames extraction:} As described in Section ~\ref{sec:frame}, frames are sampled as the representative ones from each walking cycle for feature learning. To study the influence of number of frames to re-id, experiments were carried out respectively with different number of frames (1 to 10 frames) sampled at equal intervals within each walking cycle. Note that the parameters of the CNNs are shared across all frames, which means the descriptions of all frames are generated by the same feature-extraction network.
\begin{table}[htb]
  \centering
  \caption{Re-id performance with different feature pooling strategies}
  \label{tab:pooling}
  \begin{threeparttable}
    \begin{tabular}{c|c|c|c} \hline
    Dataset & \multicolumn{1}{c|}{iLIDS} & \multicolumn{1}{c|}{PRID} & \multicolumn{1}{c}{SDU} \\ \hline
    Pooling  & R-1  & R-1  & R-1   \\ \hline
Max-pooling& {\bf60.2}    & {\bf83.3}    & {\bf89.3}   \\
Average-pooling   & 55.3  & 81.7    & 88.0    \\
Without pooling\tnote{1}  & 50.9      & 79.7     & 88.7    \\
    \hline
    \end{tabular}

 \begin{tablenotes}
        \footnotesize
\item[1] Using the feature maps of the the first frame for description.
      \end{tablenotes}
\end{threeparttable}
\end{table}

The results are given in Fig.~\ref{fig:viewN} and Table~\ref{tab:views}. Roughly speaking, the performance of using different number of sampled frames is comparable, which demonstrates our claim that it is unnecessary to use all frames for video re-id. For all datasets, four-frame sampling is the best choice and produced the best results. This is because the four frames are sampled at the maximum, minimum and middle of them in a circle, and thus contains all distinctive walking poses as illustrated in Fig. ~\ref{fig:walkingcycle}. In most cases, one or two frames gives poor results as it is too short to offer sufficient information. It is interesting to see that adding more frames does not help, because the information for identification is already redundant and more outliers may be incurred in feature learning.

To further validate the effectiveness of representative frames, experiments were conducted to compare the proposed frame sampling method to other baseline sampling methods. As shown in Tab.~\ref{tab:frames}, By dividing the sequence into $K$ walking circles, the proposed method can select the most representative frames, and performed better than randomly selection in terms of re-id accuracy. It also shows superiority over that of using all frames, which demonstrates the observation again that there is no need to extract features from all frames in video re-id.
\begin{table}[htb]
  \centering
  \caption{Performance of feature description with different layers}
  \label{tab:layers}
    \begin{tabular}{c|cc|cc|cc} \hline
    Dataset & \multicolumn{2}{c|}{iLIDS} & \multicolumn{2}{c|}{PRID} & \multicolumn{2}{c}{SDU} \\ \hline
   Layer & R-1 & R-5  & R-1 & R-5   & R-1 & R-5 \\ \hline

FC6          & 60.2  & {\bf85.1}  & {\bf83.3}  & {\bf93.3}   & 89.3  & {\bf95.3} \\
ReLU6        & {\bf60.4}  & 83.3  & 82.0  & 92.0    & 89.3&{\bf95.3}\\
FC7 &56.9&83.1&78.3&92.9&{\bf90.0}&{\bf95.3}\\
ReLU7        & 57.1&83.6&79.0&91.7& 88.7&94.7\\
Softmax  &49.3&78.2&73.7&90.0&86.0&{\bf95.3}\\
    \hline
    \end{tabular}\mytabspace
      \vspace{-5mm}
\end{table}

\paragraph{Feature pooling settings:} In this work, each sampled frame of the walking cycles is fed into the proposed network for feature extraction separately and aggregated at the feature pooling layer as shown in Fig.~\ref{fig:cnn}. Hence, feature pooling layer has an important impact on the feature aggregation as well as the final identification. As mentioned in Section~\ref{sec:pooling}, there are mainly two kinds of pooling strategies, max-pooling and average-pooling.

As shown in the Table~\ref{tab:pooling}, experiments were carried out to test the performance of max-pooling and average-pooling for the proposed network. The performance of using the features of the first frame (i.e., without pooling) is also provided as baseline for comparison. Apparently, accumulating features from multiple frames via pooling provides gains for re-id. Also, max-pooling shows superiority over average-pooling. This is unsurprising because average-pooling is usually employed in the cases within which all the input frames are considered equally important, while max-pooling cares more about the strongest (distinctive) information of each frame.
%
%

Besides, we have also considered different locations to place the feature pooling layer in the proposed network. The performance does not change much when pooling is set at the layer after $Conv_5$, however decreases evidently among the first few layers before $Conv_5$. Generally, we observed that pooling between $Conv_5$ and $FC_6$ works slightly better, and thus was used for all experiments.


\paragraph{Description layer evaluation:}
Table ~\ref{tab:layers} shows the re-id results with different layers for feature description after feature pooling. Roughly, the former layers performs better than the latter layers, and $FC_6$ yields the highest accuracy in most cases. Besides, as we illustrated before, the $ReLU$ layer is also considered as it serves as the neuron activation function after each $FC$ layer. However, there only exist slight difference between each $FC$ layer and its corresponding $ReLU$ layer. Hence, the $FC_6$ layer is used for the feature description in this work.

\paragraph{Other base networks:} Experiments have been carried out with some other base networks, such as VGG-19, Caffenet and Resnet-50. As shown in Tab.~\ref{tab:network}, VGG-M (proposed method) is better than the other networks on the provided datasets.
\begin{table}[htb]
  \centering
  \caption{Comparison to different base networks}
  \label{tab:network}
    \begin{tabular}{c|cc|cc|cc} \hline
    Dataset & \multicolumn{2}{c|}{iLIDS} & \multicolumn{2}{c|}{PRID} & \multicolumn{2}{c}{SDU} \\ \hline
   Network & R-1 & R-5  & R-1 & R-5   & R-1 & R-5 \\ \hline
VGG-19     &43.3 &70.0 &70.0 &91.0 &87.3 &93.7 \\
Caffenet   &46.7 &74.7 &67.0 &93.0 &88.7 &93.7 \\
Resnet-50   &56.7 &{\bf85.3} &74.0 &91.0 &85.3 &93.1 \\
VGG-M      &{\bf60.2} &85.1 &{\bf83.3} &{\bf93.3} &{\bf89.3} &{\bf95.3}\\

    \hline
    \end{tabular}\mytabspace
      \vspace{-5mm}
\end{table}

\begin{figure}[htbp]
\vspace{-5mm}
\includegraphics[width=\linewidth]{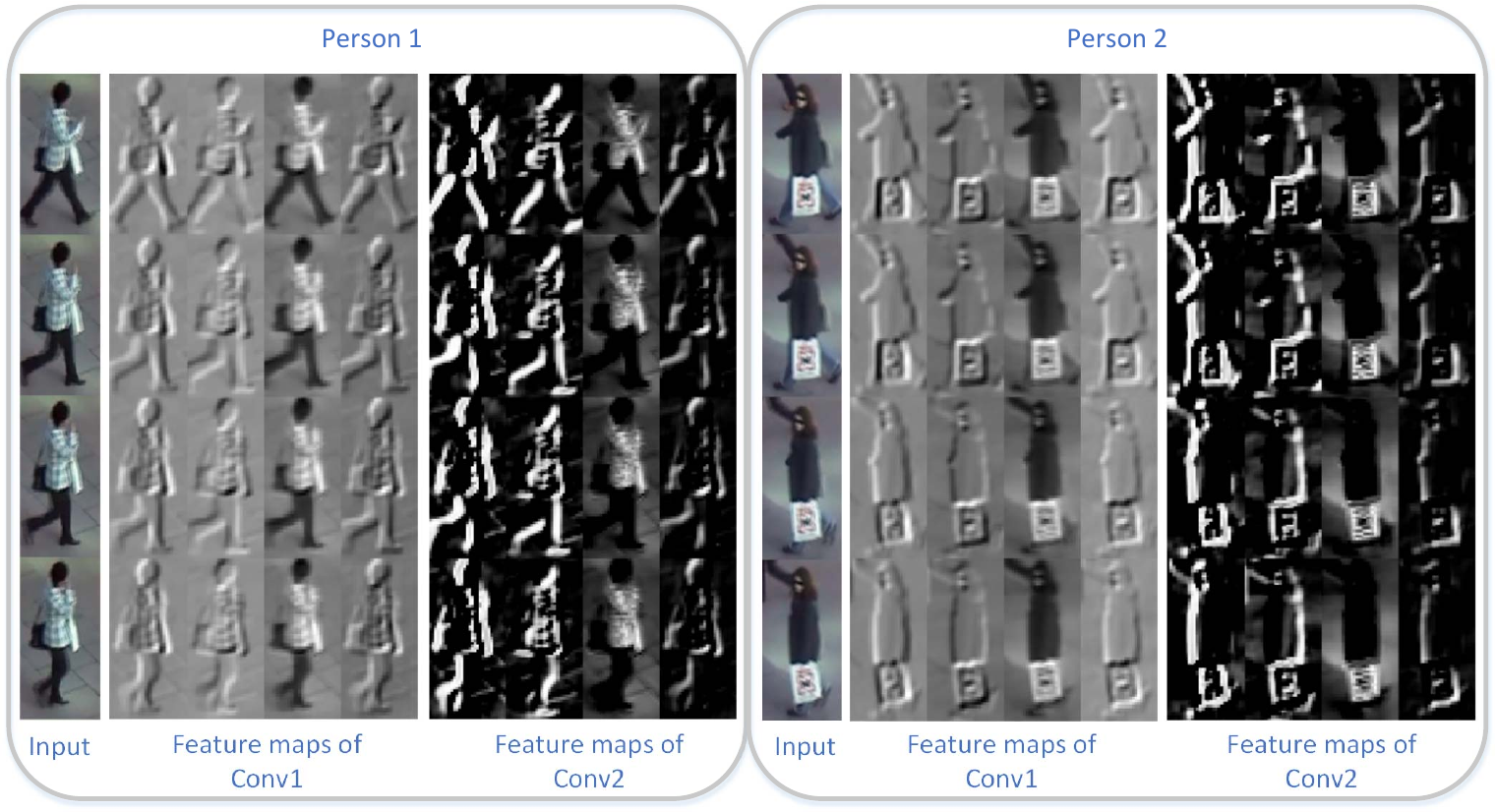}
\caption{Visualization of the learned feature maps.}\label{fig:feats2}
\end{figure}

\paragraph{Feature map visualization:} To validate the proposed appearance representation, we intend to visualize the learned intermediate features. Fig.~\ref{fig:feats2} shows two examples, and each of them presents some feature maps produced by $Conv_1$ and $Conv_2$. As expected, most representative features, including silhouettes and distinctive appearance like clothes and bags, can be captured by the proposed learning model. The feature pooling layer is capable of combining all representative features learned at different walking states (frames) into a joint representation for more effective person identification.
\subsection{Results of distance learning}
\label{sec:kissme}
\begin{table}[htb]
  \centering
  \caption{Comparison to different distance metric learning methods}
  \label{tab:distance}
    \begin{tabular}{c|c|c|c|c} \hline
    \multicolumn{2}{c|}{Dataset} & \multicolumn{1}{c|}{iLIDS} & \multicolumn{1}{c|}{PRID} & \multicolumn{1}{c}{SDU} \\ \hline
   \multicolumn{2}{c|}{Distance metric learning} & R-1  & R-1  & R-1  \\ \hline
 \multirow{2}{*}{\tabincell{c}{KISSME\\ \cite{KoestingerCVPR12}}} &$d_{min}$ &52.0  &82.3   &85.3 \\
&$d_{avg}$ &{\bf60.2} &{\bf83.3} &89.3 \\
\hline
 \multirow{2}{*}{\tabincell{c}{LFDA\\ \cite{PedagadiCVPR13}}}  &$d_{min}$ &49.6&72.0  &80.7  \\
 &$d_{avg}$ &57.3  &76.7 &86.0 \\
\hline
 \multirow{2}{*}{\tabincell{c}{XQDA\\ \cite{liao2015person}}}  &$d_{min}$ &49.3&74.3  &84.7  \\
 &$d_{avg}$ &59.1 &80.3 &{\bf90.7} \\
    \hline
    \end{tabular}
\end{table}
\paragraph{Metric learning evaluation:} In this experiment, we combine the proposed network with different supervised distance metric learning methods such as KISSME \cite{KoestingerCVPR12}, Local Fisher Discriminant Analysis (LFDA) \cite{PedagadiCVPR13} and Cross-view Quadratic Discriminant Analysis(XQDA) \cite{liao2015person}. As shown in Table~\ref{tab:distance}, among the three methods, KISSME performed best in most cases and thus was chosen as the default method for distance metric learning.

\paragraph{Distance measure evaluation:} Table~\ref{tab:distance} also gives the testing of classifiers with different distance measures: minimum distance in Eq.(\ref{measure1}) and average distance in Eq.(\ref{measure2}). It is observed that, the classifier with average distance $d_{avg}$ performs better than the one with minimum distance measure $d_{min}$, especially on the iLIDS-VID dataset. This is mainly because the average classifier is more resilient to noise caused by occlusion and light changing, which happens more frequently in the first dataset. 
%


\paragraph{Dimension reduction evaluation:} Appropriate dimension reduction not only help preserve discriminative information, but also help filter out the noises in features. The effect of dimension reduction using PCA is studied in Table~\ref{tab:pca}. The optimal performance is obtained with the dimension reduced to 100 using PCA.
\begin{table}[htb]
  \centering
  \caption{Evaluation of different PCA dimension reduction}
  \label{tab:pca}
    \begin{tabular}{c|c|c|c} \hline
    Dataset & \multicolumn{1}{c|}{iLIDS-VID} & \multicolumn{1}{c|}{PRID 2011} & \multicolumn{1}{c}{SDU-VID} \\ \hline
    PCA  & R-1  & R-1 & R-1  \\ \hline
50    &52.9  &79.0 &86.0 \\
100   &{\bf60.2} &{\bf83.3}&89.3 \\
150   &58.9 &81.3 &88.7 \\
200   &56.0 &77.0 &{\bf90.7}\\
    \hline
    \end{tabular}\mytabspace
      \vspace{-3mm}
\end{table}

\begin{table*}[htb]
  \centering
  \caption{Comparison to state-of-the-art methods on benchmark datasets}
  \label{tab:methods}
    \begin{tabular}{c|ccc|ccc|ccc} \hline
    Dataset & \multicolumn{3}{c|}{iLIDS} & \multicolumn{3}{c|}{PRID} & \multicolumn{3}{c}{SDU} \\ \hline
    Methods  & R-1 & R-5 & R-20 & R-1 & R-5 & R-20 & R-1 & R-5 & R-20 \\ \hline
GEI+RSVM \cite{MartinECCV12}   &2.8&13.1&34.5&- &- &- &- &- &-\\
HOG3D+DVR \cite{WangECCV14}     &23.3&42.4&68.4& 28.9&55.3&82.8&-&-&-   \\
Color+LFDA \cite{PedagadiCVPR13} &28.0&55.3&88.0& 43.0&73.1&90.3&-&-&-  \\
STA \cite{liu2015spatio}     &44.3&71.7&91.7& 64.1&87.3&92.0&73.3&92.7&96.0 \\
RNN \cite{mclaughlinrecurrent}   &50.0&76.0&94.0&65.0&90.0&97.0&75.0&86.7&90.8 \\
RNN+OF \cite{mclaughlinrecurrent} &58.0&84.0 &{\bf96.0} &70.0 &90.0 &97.0 &-&-&- \\
CNN+XQDA \cite{zheng2016mars}     &53.0&81.4&95.1&77.3& {\bf93.5}&{\bf99.3}&-&-&-  \\
Ours    &{\bf60.2} &{\bf85.1} &94.2 &{\bf83.3} &93.3 &96.7 &{\bf89.3} &{\bf95.3} &{\bf98.5}  \\
    \hline
    \end{tabular}
\end{table*}


\subsection{Comparison to state-of-the-art}
In this section, we compare the performance of the proposed method to existing video-based re-id approaches as shown Table ~\ref{tab:methods}. It can be observed that the proposed algorithm achieved state-of-the-art performance on the benchmark public datasets. For iLIDS-VID, our algorithm outperforms the second best one: RNN+OF \cite{mclaughlinrecurrent} by $2.2\%$. For PRID 2011, our algorithm outperforms the second best one: CNN+XQDA \cite{zheng2016mars} by $6\%$. For SDU-VID, only the results of STA \cite{liu2015spatio} and RNN \cite{mclaughlinrecurrent} are provided, and our method produced significant gains of $14.3\%$. It should be stressed that the above methods takes all the frames as input and the performance mostly rely on the motion features extracted using hybrid tools, e.g., RNN \cite{mclaughlinrecurrent}, optical flow \cite{lucas1981iterative}, HOG3D \cite{KlaserBMVC08} and GEI \cite{HanTPAMI06}. In contrast, the propsed method yileds superior results by pooling the image features from only a few frames.

\subsection{Limitations and discussions}
\begin{figure}[htbp]
\includegraphics[width=\linewidth]{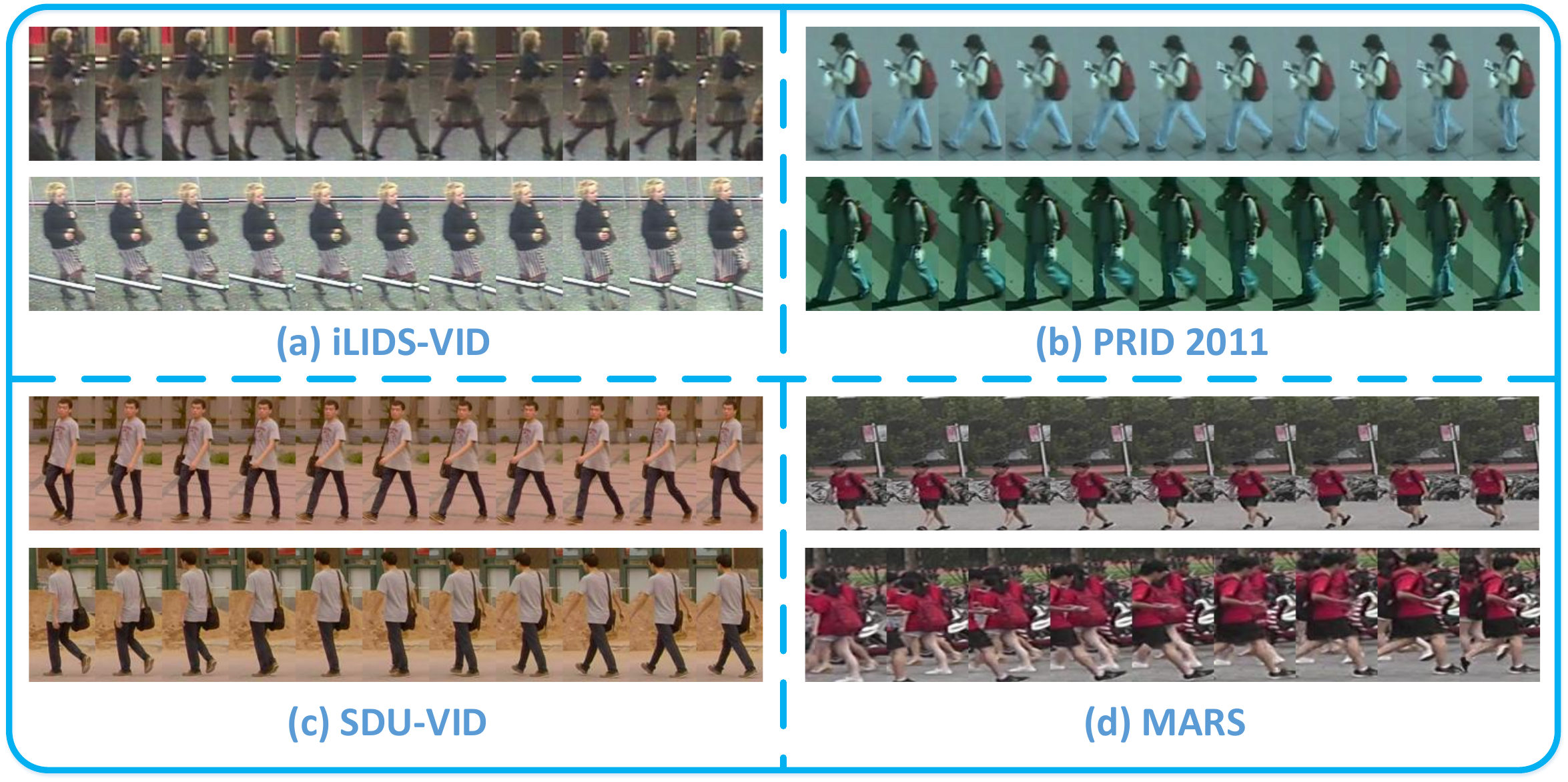}
\caption{(a)Sample sequences in iLIDS-VID. (b)Sample sequences in PRID 2011. (c)Sample sequences in SDU-VID. (d)Sample sequences in MARS.}\label{fig:datasets}\myfigspace
\end{figure}
Besides iLIDS-VID \cite{WangECCV14}, PRID 2011 \cite{HirzerSCIA11} and SDU-VID \cite{liu2015spatio}, a new dataset: MARS \cite{zheng2016mars} was developed recently, and differs much from the other three datasets. As shown in Fig.~\ref{fig:datasets}, the pedestrians of the earlier publicly datasets were mostly captured from sideview, while the camera viewpoints and poses in the MARS vary greatly, and the length of tracklets is much smaller. Besides, as shown in Fig.~\ref{fig:datasets}(d), since the pedestrian detection and tracking were performed automatically, the quality of the cropped pedestrian is poor. As mentioned by the authors, quite a number of distractor tracklets were produced by false detection or tracking results. All these issues pose great difficulty for our method to extract walking circle and representative frames. Also, the feature pooling is also fragile to the ambiguity incurred by the large portion of background or other scene object. Since extracting the representative frame by walking circle is intractable, we split each tracklet into half and randomly select four frames as the representative ones for feature learning and pooling. The results are given in Table \ref{tab:mars}. Our method is inferior to CNN+XQDA \cite{zheng2016mars} in this case. This is reasonable as CNN+XQDA takes all frames for different sequences. Without the representative frames, our method can only process a constant number of frames sampled randomly (e.g., four frames) and thus is more sensitive to the above issues in MARS. The above limitations are shared among most methods as shown in Tab.~\ref{tab:mars}.

\begin{table}[htb]
  \centering
  \caption{Testing on the MARS dataset}
  \label{tab:mars}
    \begin{tabular}{c|ccc} \hline
    Dataset & \multicolumn{3}{c}{MARS}\\ \hline
    Methods  & R-1 & R-5 & R-20  \\ \hline
HistLBP+XQDA  &18.6 &33.0 &45.9 \\
BoW+KISSME  &30.6 &46.2 &59.2  \\
SDALF+DVR   &4.1 &12.3 &25.1 \\
HOG3D+KISSME  &2.6 &6.4 &12.4 \\
GEI+KISSME &1.2 &2.8 &7.4\\
CNN+XQDA   &{\bf65.3} &{\bf82.0} &{\bf89.0}  \\
Ours    &55.5 &70.2 &80.2 \\
    \hline
    \end{tabular}\mytabspace
      \vspace{-3mm}
\end{table}
\section{Conclusions}
\label{sec:conclusions}
In this paper, we presented a novel video-based person re-id framework based on deep CNNs. Unlike the previous work focusing on extracting the motion cues, the efforts were spent on extracting compact but discriminative appearance feature from typical frames of a video sequence. The proposed appearance model was built with a deep CNN architecture incorporated with feature pooling. Extensive experimental results on benchmark datasets confirmed the superiority of the proposed appearance model for video-based re-id.



{\small
\bibliographystyle{ieee}
\bibliography{referencefile}
}

\end{document}